# The FeatureCloud AI Store for Federated Learning in Biomedicine and Beyond

Julian Matschinske[1], Julian Späth[1], Reza Nasirigerdeh[2], Reihaneh Torkzadehmahani[2], Anne Hartebrodt[3], Balázs Orbán[4], Sándor Fejér[4], Olga Zolotareva[5], Mohammad Bakhtiari[1], Béla Bihari[4], Marcus Bloice[6], Nina C Donner[7], Walid Fdhila[8], Tobias Frisch[3], Anne-Christin Hauschild[9], Dominik Heider[9], Andreas Holzinger[6], Walter Hötzendorfer[10], Jan Hospes[10], Tim Kacprowski[11,12], Markus Kastelitz[10], Markus List[5], Rudolf Mayer[8], Mónika Moga[4], Heimo Müller[6], Anastasia Pustozerova[8], Richard Röttger[3], Anna Saranti[6], Harald HHW Schmidt[13], Christof Tschohl[10], Nina K Wenke[1], Jan Baumbach[1,3]

1. Chair of Computational Systems Biology, University of Hamburg, Hamburg, Germany
2. Institute of AI in Medicine and Healthcare, Technical University of Munich, Munich, Germany
3. Institute of Mathematics and Computer Science, University of Southern Denmark, Odense, Denmark
4. Gnome Design SRL, Sfântu Gheorghe, Romania
5. Chair of Experimental Bioinformatics, Technical University of Munich, Freising, Germany
6. Institute for Medical Informatics, Statistics and Documentation, Medical University of Graz, Graz, Austria
7. concentris research management gmbh, Fürstenfeldbruck, Germany
8. SBA Research Gemeinnützige GmbH, Vienna, Austria
9. Department of Mathematics and Computer Science, Philipps-University of Marburg, Marburg, Germany
10. Research Institute AG & Co KG, Vienna, Austria
11. Division Data Science in Biomedicine, Peter L. Reichertz Institute for Medical Informatics of TU Braunschweig and Hannover Medical School, Brunswick, Germany
12. Braunschweig Integrated Centre of Systems Biology (BRICS), Brunswick, Germany
13. Department of Pharmacology and Personalised Medicine, MeHNS, FHML, Maastricht University, Maastricht, The Netherlands.

# Abstract

Machine Learning (ML) and Artificial Intelligence (AI) have shown promising results in many areas and are driven by the increasing amount of available data. However, this data is often distributed across different institutions and cannot be shared due to privacy concerns. Privacy-preserving methods, such as Federated Learning (FL), allow for training ML models without sharing sensitive data, but their implementation is time-consuming and requires advanced programming skills. Here, we present the FeatureCloud AI Store for FL as an all-in-one platform for biomedical research and other applications. It removes large parts of this complexity for developers and end-users by providing an extensible AI Store with a collection of ready-to-use apps. We show that the federated apps produce similar results to centralized ML, scale well for a typical number of collaborators and can be combined with Secure Multiparty Computation (SMPC), thereby making FL algorithms safely and easily applicable in biomedical and clinical environments.

# Main

Machine learning (ML) has risen tremendously in popularity during the last decade. Enabled by increased computational capacities and novel concepts, it has been used to gain new insights in various fields, including biomedicine[1,2]. As a rule, the quality of ML models improves with the size of the available data. However, data is often scattered across multiple facilities, and privacy regulations such as the European General Data Protection Regulation (GDPR) often restrict data sharing, rendering large-scale, centralized ML infeasible. Particularly in biomedicine, the large-scale collection of molecular and clinical data is becoming ubiquitous with successful applications of ML in diagnostics[3] or drug discovery[4]. However, privacy concerns hinder even faster advances and sometimes render the usage of ML impossible altogether due to small sample sizes of the individual datasets available, e.g. in case of rare diseases. Federated Learning (FL) has been suggested as a method to overcome these challenges. The key advantage of FL is keeping the data on the data holders machine and exchanging only model parameters. Thereby, a model using all the data can be trained and exploited for research and diagnostics without the risk of disclosing any primary data.

FL is already being used in privacy-aware contexts, such as smartphone apps, where millions of users collectively improve models for the autocomplete functionality[5]. But despite its demonstrated feasibility, obvious advantages, and huge potential, FL still lacks adoption in many areas, including collaborative research.

Other privacy-preserving techniques like homomorphic encryption (HE) have been suggested and successfully employed to address these concerns[6]. HE allows for ML on encrypted data. However, these techniques are computationally expensive and require profound changes to the original ML algorithm. In contrast, FL is a relatively simple and efficient approach, compatible with most ML algorithms. Recent work demonstrates that federated implementations of various ML algorithms yield comparable or even identical results while maintaining a sufficient level of privacy[7–9]. FL alone cannot always fulfill strict privacy requirements[10]. However, it can be combined with other techniques, such as secure multiparty computation (SMPC)[11] or differential privacy (DP)[12], to achieve a definable and provable level of privacy.

Traditional ML algorithms can be directly applied to centralized datasets and usually require no changes since many off-the-shelf implementations exist. Tuning of hyper-parameters or finding suitable architectures for neural networks can be done by a single person on a single machine. In contrast, while its theoretical design is easy to grasp, FL, in practice, requires a complex software system instead of just a single script. To facilitate and coordinate the exchange of ML parameters, this system needs to potentially operate on dozens of machines with different operating systems, varying network connectivity, and computation capacity. Prior to any computation, an algorithm needs to be conceptually divided into steps that have to be run on the participants' machines and parameter aggregation steps that need to be performed by an aggregating party. Similar to centralized ML, collaborative ML, individual pre-processing of a dataset is still required to transform the data into a common format compatible with the algorithm. The development of such distributed systems is complicated and time-consuming. To facilitate the use of FL in the research community, we developed FeatureCloud, an all-in-one solution covering these steps in a generic way (see Figure 1). FeatureCloud facilitates the development and deployment of federated tools and lowers the technical barrier for end-users.

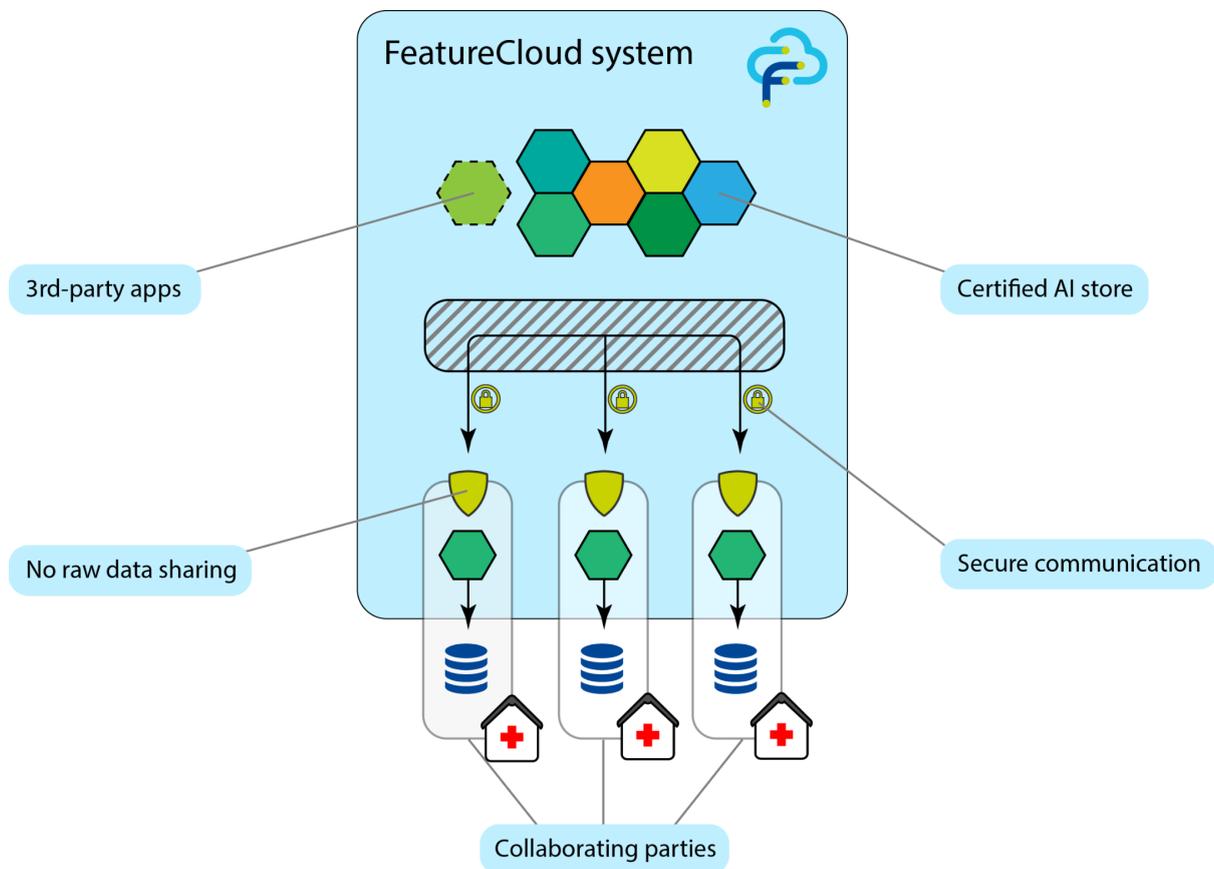

*Figure 1. Outline of the FeatureCloud system.* *Medical institutions collaborate in a federated study while all primary or raw data remains at its original location. FeatureCloud handles distribution, execution, and communication of certified AI apps from the FeatureCloud AI Store.*

Several frameworks for FL have been developed recently. They often focus on DL models, which are hard to interpret and thus not well suited to justify medical decisions. Two different categories can be distinguished: 1) backend-only and 2) all-in-one approaches. Backend frameworks provide developers with methods to simplify the implementation of federated and privacy-aware ML algorithms[13–16]. They are limited to users with a background in software development or programming experience. Such know-how can usually not be expected from clinical experts, and therefore restricts usability considerably. All-in-one frameworks bring privacy-aware analyses to users without in-depth programming skills by providing a graphical user interface (GUI)[17–20].

Each existing framework has its benefits and can be useful in specific scenarios. However, only few of them are suited for practical application in clinical environments. Notably, to the best of our knowledge, there is no comprehensive open-source solution that combines the crucial tasks necessary for practically using FL in a specific environment: development, distribution, and execution of federated AI methods. While some platforms attempt to facilitate the federated training of models for the end-user, the resulting models are either not collected and shared with all collaborating parties, their application programming interface (API) is not open to developers, restricting end-users to predefined models, or novel apps from external developers cannot be shared easily. This is a huge disadvantage since the area of AI is rapidly evolving and algorithms are enhanced frequently.

The FeatureCloud AI Store accelerates all of the steps involved in FL, specifically the development of federated algorithms by providing an open API, as well as deployment, distribution, and usage of algorithms through configurable workflows. The AI Store is open to external developers, allowing them to add their own federated apps. FeatureCloud Apps are publicly available to end-users and can be executed without requiring any programming skills. This makes them available to researchers or doctors, for instance in academic institutions or medical facilities, who can then use their data together with other parties for collaborative FL. Typical app developers are expected to be end-users with an additional programming background who develop AIs for general research purposes. Although FeatureCloud was primarily designed for biomedical and clinical research, it may ultimately be used in any ML-affine domain.

# Results

## The FeatureCloud AI Store

The FeatureCloud AI Store provides an intuitive and user-friendly interface for both biomedical researchers and developers. It already contains a variety of apps and displays basic information about them, including short descriptions, keywords, end-user ratings, and certification status. Users can easily find apps of interest *via* a textual search and filter them by type (pre-processing, analysis, evaluation) and their privacy-preserving techniques (FL, DP, HE). End-users can review the apps and provide valuable feedback on the app quality. The app details page displays a method summary, app description, user reviews, developer name, and contact details to report bugs. Each app provides either a graphical frontend or a simple configuration file to set app parameters and adapt them to different contexts. This reduces technical details and makesapps user-friendly for end-users independent of their background.

The AI Store comes with a broad selection of popular machine learning models, listed in Table 1. The apps are categorized into preprocessing, analysis, and evaluation. Some analysis apps, such as linear regression (LR) and random forest (RF) are generic and suitable for different data types and a wide range of scenarios. Such apps can easily be assembled into a workflow with generic preprocessing and evaluation apps, e.g. for standardization and evaluation of classifiers. Other apps, such as the Chi-squared test for Genome-Wide Association Studies (GWAS), integrate all necessary steps of a certain application-specific workflow and do not require combination with other apps.

| Name | Category | Features or domain |
| --- | --- | --- |
| Cross-validation | Pre-processing | Creates k-fold splits |

| Normalization | Pre-processing | Standardization, min-max and max-abs scaling |
| --- | --- | --- |
| Feature Selection | Pre-processing | Selection based on Pearson correlation |
| Linear Regression | Analysis | Trains a federated linear regression model |
| Logistic Regression | Analysis | Trains a federated logistic classification model |
| Random Forest | Analysis | Trains a merged random forest for regression or classification |
| Support Vector Machine | Analysis | Linear SVM and weighted ensemble of SVMs for classification |
| Classifier Evaluation | Evaluation | Calculates accuracy, precision, recall, F-score and Matthew's correlation coefficient |
| Regression Model Evaluation | Evaluation | Calculates mean absolute error, max error, (root) mean squared error and median absolute error |
| Kaplan-Meier Estimator | All-in-one | Estimates the survival function of time-to-event data |
| Nelson-Aalen Estimator | All-in-one | Estimating the cumulative hazard ratio of time-to-event data |
| Cox Proportional Hazard | All-in-one | Multivariate regression for time-to-event data |
| Random Survival Forest | All-in-one | Random forest for time-to-event data |
| AdaBoost | All-in-one | Boosting, classification using random forest |
| Chi-squared test for GWAS | All-in-one | Perform a Genome-Wide Association Study using the Chi-squared test |

*Table 1. List of apps in the FeatureCloud AI Store.* *The growing list of apps available in the AI Store covers pre-processing, analysis and evaluation. All-in-one apps cover the whole workflow for a more specific domain and can be executed without other apps.*

## App development and certification process

We do not want to restrict end-users to the current selection of apps, so FeatureCloud invites external developers to implement their own federated apps and publish them in our AI Store. A FeatureCloud app is a program, isolated inside a Docker container, that communicates with other instances through the FeatureCloud API. Therefore, the API description is openly available (see Supplementary File B). Several templates and example apps are provided to further facilitate the implementation by directly explaining the API with code.

Besides the AI Store and the API, FeatureCloud provides tools accelerating the development of federated applications. When developing a new federated method, app developers can directly start with the federation of the AI logic by using an existing template. To verify that the API has been implemented correctly, a simulation tool aids the developer to test their app before publishing. For each test run, it allows for specifying the number of participants, test data and communication channels and subsequently starts the according instances, simulating a real-world execution on multiple machines. During the test run, it shows logs and results for each participant and the network traffic to monitor the execution and identify bugs and potential communication bottlenecks.

After the development phase, apps can be published in the FeatureCloud AI Store. Developers need to fill a form prompting all relevant information about the app, which is subsequently displayed to the end-users and used for the search and filter functions. After that, they can push their Docker image to the Docker registry of the FeatureCloud platform. For end-users collaborating with the developer, who explicitly enable uncertified apps, it is already usable by and can be tested in a real-world scenario. For other end-users, we enforce a certification process intended to block malicious apps and maintain high privacy standards in the AI Store. To this end, the developer needs to provide necessary documentation and details about the implemented privacy mechanism. Furthermore, the source code of the app must be accessible so that the app can be exhaustively tested and vetted by the FeatureCloud team for possible privacy leaks. When the certification process has been successfully completed by a member of the FeatureCloud consortium according to a defined checklist (see Supplementary File D), the app will be displayed in the AI Store and can be used by all end-users. If the certification process was unsuccessful, the developer is notified and is requested to address the issues that have been raised. Upon each update of an application, a new certification procedure is triggered.

## Secure Multiparty Computation (SMPC)

For the initial release of FeatureCloud, the API has been kept general (see Supplementary File B) to impose as few restrictions as possible on the supported algorithms. However, while FL significantly increases privacy, it can still leak information to the coordinator, who can see all individual models before aggregating them. The same applies to potential security breaches where the network traffic is intercepted. Local updates of the model based on a previously distributed global model may potentially reveal information about the primary data. Secure Multiparty Computation (SMPC) has been suggested as a solution[11] to this issue. An adapted SMPC implementation has thus been integrated into FeatureCloud and is provided as an app template for the API. It involves an additional iteration before each aggregation step, increasing the overall runtime by 2-fold in most cases, while substantially decreasing the risk of network interception and removing the need to trust the aggregating party more than all other participants.

## Cross-institutional analyses

Before collectively running a federated *workflow*, all collaborating partners (aka *participants*) have to download and start the client-side FeatureCloud Controller on their machines. It only requires Docker, which is freely available for all major operating systems. Users also need to create an account on the FeatureCloud website (www.featurecloud.ai), which serves as a web frontend and is used to coordinate the FeatureCloud system (see Methods and Supplementary File A for details on the architecture). When these requirements are met, all participants are ready to run apps from the AI Store. Each collaborative execution of apps is organized into so-called *projects* on the web frontend. They contain a description of the planned analysis, connect the collaborating partners by allowing invited participants to join, and show the current status of the workflow (see also Supplementary Figure S1).

Workflows are composed of one or multiple apps that are to be executed consecutively. Each app produces intermediate results that serve as input for the consecutive app. Intermediate results are kept on the respective machines and are not shared with other participants. The last app produces the final results, which are then shared with all project participants. During the execution of a workflow, its progress can be monitored on the FeatureCloud website, showing the current stage, computation progress, and intermediate results from each individual app. Apps can provide their own frontend user interface, allowing for user interaction if necessary and for showing specific reports. Users can monitor app logs and react in case something unexpected happens (e.g. stop and re-run the workflow with other data or a different configuration). When the last app in the workflow successfully completes its computation, the final results are shared automatically with all project participants. Intermediate results and app logs remain available on the local machines to allow for later verification. For instance, the results may include a report showing the effectiveness of the trained model and the model itself. The latter can subsequently be used outside of FeatureCloud as well. If a project fails, e.g. because a participant drops out, it can be restarted easily after the cause has been dealt with. During the entire process, no programming knowledge or command-line interaction is required, making the system especially suited for medical personnel without a technical education.

## Evaluation

To evaluate the applicability of FeatureCloud in practice, multiple workflows operating on different datasets were composed. Each workflow consists of a cross-validation (CV) app, a standardization app, a model training app, and a final evaluation app (see Figure 2). The individual apps are datatype-agnostic and suited for various applications. For classification analyses, they were evaluated on the Indian Liver Patient Dataset[21] (ILPD) with 579 samples and 10 features and the Cancer Genome Atlas Breast Invasive Carcinoma[22] (BRCA) dataset with 569 samples and 20 features. For regression analyses, they were evaluated on the Diabetes[23] dataset with 442 samples and 10 features and the Boston[24] house prices dataset with 506 samples and 13 features, both provided by scikit-learn[25]. For each workflow, we investigated the overall effectiveness using 10-fold CV for 5 participants with an uneven distribution of data, where participant 1, participants 2-3 and 4-5 each have 10%, 15% and 30% of the samples, respectively. We also investigated the scalability with respect to runtime and network traffic for 2 to 8 participants.

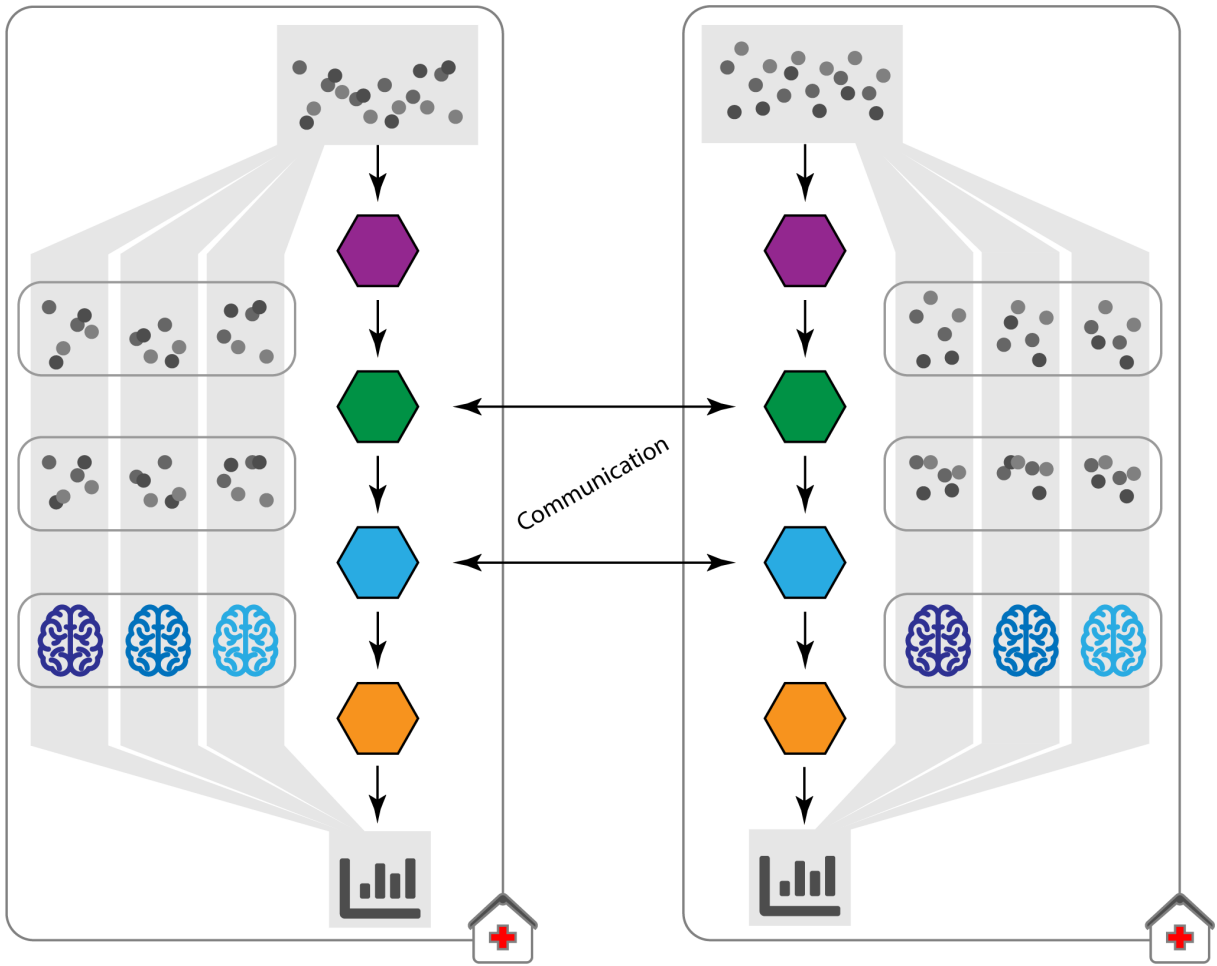

*Figure 2. Workflow structure used for evaluation. The first app (purple) creates splits for cross-valuation. All following apps perform their tasks on each split individually, in a federated fashion, only transmitting model parameters. The grey dots represent intermediate training/test data. The second app (green) performs normalization and the third (blue) trains the models, generating a global model. The global model is evaluated in the evaluation app (orange). The evaluation results are finally aggregated to obtain an evaluation report, based on the initial CV splits.*

### Performance

Previous studies have shown that FL can achieve similar performance to centralized learning in many scenarios[9,26,27]. To verify the approach used in FeatureCloud, we compared the performance of four federated FeatureCloud apps integrated into an ML workflow to their corresponding centralized scikit-learn[25] models. The results are shown in Figure 3. For logistic regression and linear regression, the FeatureCloud workflow achieved performance identical to scikit-learn, which is consistent with previous results of federated linear and logistic regression applications[75,7]. For the random forest regression and classification models, a similar performance was achieved. Due to the simple aggregation method that combines the local trees into one global tree, identical results were not obtained nor expected. Due to the bootstrapping mechanism and its attached randomness, the federated random forest sometimes performs slightly better than the centralized approach.

Furthermore, we compared the federated models to the individual models trained and evaluated at each individual participant (also 10-fold CV). Here we distinguish between central evaluation of the models on the overall test splits (central test data), identically to the test splits for the centralized and federated models, and local evaluation on the local test splits only (local test data). As shown in Figure 3, central evaluation performance varies widely but is on average worse than for the federated models. For classification, also the local evaluation performs worse than for the federated models. For the regression models however, the centrally evaluated models of the individual participants even outperform the centralized model in some cases. Nevertheless, compared to the central test data, it is obvious that these models did not generalize well and were only performing well at the individual participants with a very small test set. This can be very deceptive, as in this case even the 10-fold CV cannot be trusted. This highlights the effectiveness of FL, as these models make use of more train and test data resulting in more generalized models.

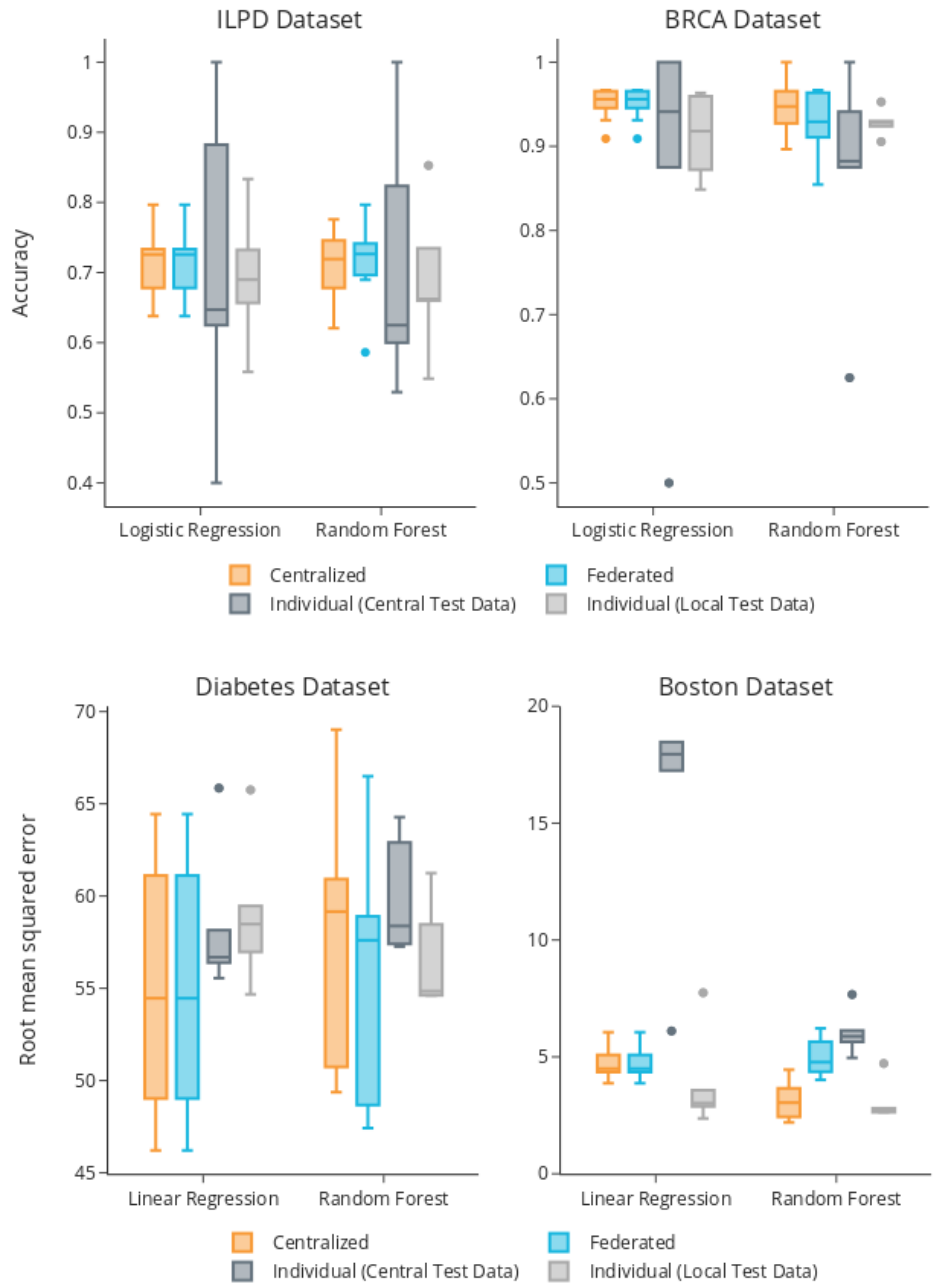

*Figure 3. Performance evaluation of federated AI methods. The boxplots show the results of a 10-fold CV for the different classification and regression models and datasets in multiple settings. The centralized results are shown in orange, the corresponding federated results in blue and the individual results obtained locally at each participant in grey. Each model was evaluated on the entire test set (dark grey) like the centralized and federated models, and on the individual (local) parts of the test set (light grey). The federated logistic and linear regressions perform identically to their centralized versions and the federated random forest performs similar to its centralized version.*

### Runtime and network traffic

Multiple executions with varying numbers of clients were performed to assess the scalability of the FeatureCloud platform and the federated methods. Random forest and linear regression classifiers were chosen as iterative and non-iterative methods, and both applied to the ILPD. Both were tested with 2, 4, 6 and 8 clients and the same number of samples to ensure comparability across the executions. To investigate the impact of network bandwidth on the runtime, all executions were performed on a normal and throttled internet connection with a maximum transmission of 100 kB/s.

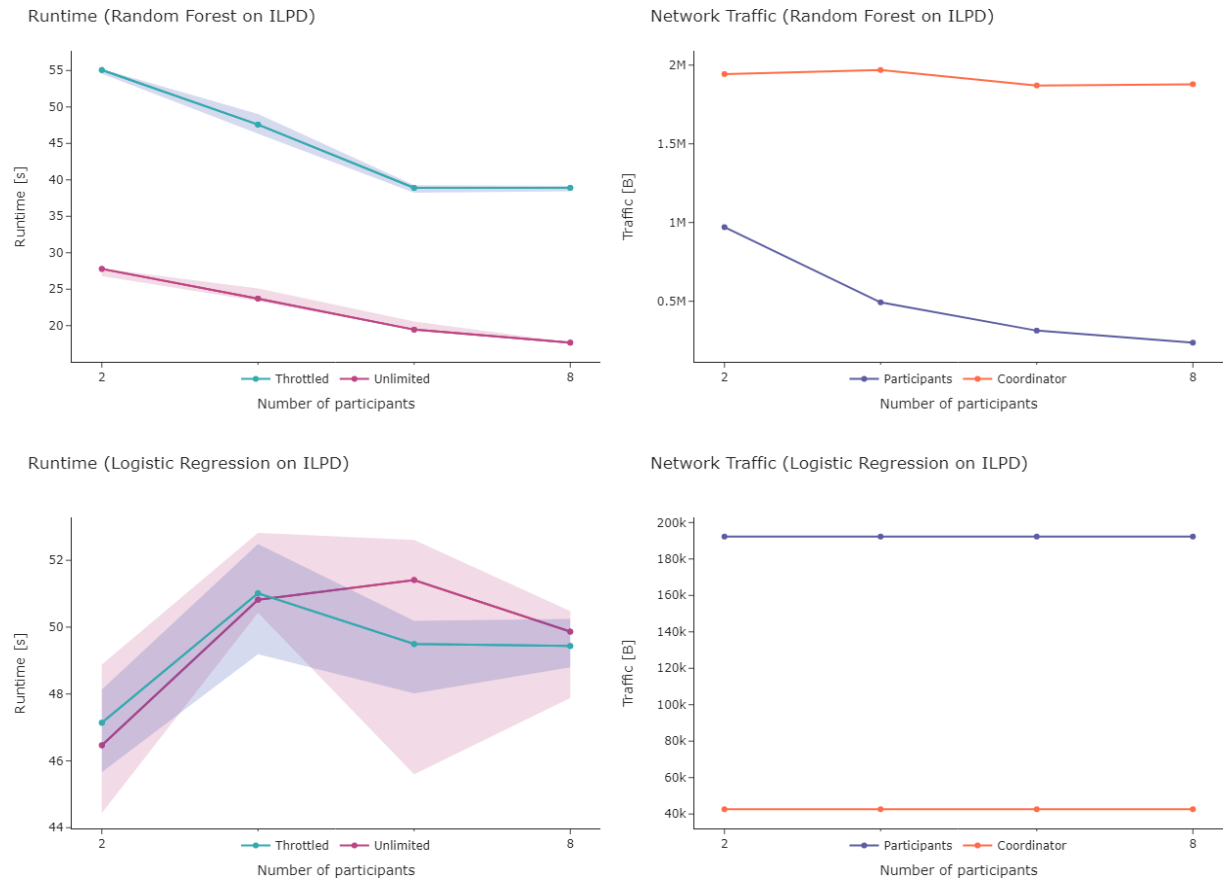

*Figure 4. Runtime and network traffic.* The left plots show runtime for unlimited and throttled connections, the right plots show network traffic for coordinator and participants evaluated on the Indian Liver Patient Dataset. The lines represent the median values measured over 10 executions. The areas show the 25% and 75% quartiles to illustrate variance across the executions.

Figure 4 shows that runtime mildly increases for logistic regression but decreases for RF. This is because LR models are of equal size for all clients, while the size of random forest models depends on the number of trees. In our implementation of federated RF, the global model is of fixed size (100 trees), which means that each client contributes a portion that decreases with a higher number of participants. Throttling bandwidth significantly increases runtime for RF but leaves the runtime for LR almost unaffected. This is because the transmitted data for RF is more extensive and comes in one chunk, whereas LR needs about 10 iterations, each exchanging a few parameters. The centralized versions take two to three seconds to complete for both LR and RF, implying their federated versions take 10 to 20 times longer to complete.

In this setting, an increasing number of participating parties has a weak impact on the duration of the aggregation part for these methods compared to the total runtime. The local computations happen in parallel so that an increasing number of participants does not have a big impact either. However, since the aggregation step cannot be completed before all participants have sent their models, the runtime of each aggregation step depends on the slowest participant, which poses a potential problem for large federations. FeatureCloud mainly focuses on being used in the tightly regulated environment of medical research. For this reason, there currently is no automatic "matchmaking" in place, but all participants need to join for each project actively. In this context, running an analysis with datasets of more than eight participants is still an uncommon scenario, which has not been evaluated in depth.

## Discussion

The FeatureCloud platform can be applied to practical problems in biomedicine and beyond. It is general enough to allow for a variety of ML algorithms, yet offers pre-built solutions for common use cases in the form of apps in the AI Store or app templates for developers. The concept of freely composing apps in a workflow proves to be challenging due to the necessity of a common data format, which is not always available and can reduce flexibility. The same applies to the initial data, which needs to be provided in a form processable and understandable by the desired apps.

Since FL adaptation is still in its early stages, it is necessary to understand which functionality and which types of data will be used, which ML techniques prove to be most prevalent in federated settings, and which challenges arise when using the platform. Therefore, few assumptions can be made in advance. FeatureCloud aims to keep the platform as flexible and extensible as possible and to align new functionality closely to the demand of its users. The possibility of integrating other privacy-preserving techniques, such as DP or SMPC, on the app layer of the API demonstrates the versatility of this approach. Even though the current implementation of SMPC suffers from quadratic increase in network traffic, it shows that flexible communication can be achieved through asymmetric encryption and can serve as a blueprint for similar scenarios and future developments.

The prediction performance of the FL apps is promising with some performing just as well as their centralized counterparts (linear/logistic regression, normalization) or almost as well (random forest). The computational and communication overhead is usually acceptable for ordinary FL and plays a smaller role than the additional overhead related to human-to-human coordination of federated projects. For up to eight participants, we demonstrate that the currently available apps and the platform scale well.

Given the flexibility of the AI Store, reaching from prebuilt task-centered apps, such as GWAS, to generic method-centered apps, such as random forest, we address a broad spectrum of end-users. Less experienced users without deeper methodological or statistical knowledge benefit from the ease of use of a task-centered app and advanced users can tailor a workflow to their needs. App developers, on the other hand, can easily reach a broad user base. They are incentivized to develop their apps to be compatible with existing ones (e.g., a new AI method processing data pre-processed by an existing normalization app) to maximize their utility. That way, the FeatureCloud AI Store aims to become an ecosystem for FL, driving collaborative research.

FeatureCloud envisions to be driven by an emerging community with its features being closely aligned to its needs. To this end, we plan to add a part on the website for the community to connect, engage with the FeatureCloud consortium and explore further steps together.

# Methods

The methods described in this section are divided into the general design of FL used in FeatureCloud, and how the ML methods used in the performance and runtime evaluation have been implemented as FeatureCloud apps.

## Federated learning (FL)

FL generally involves two possibly alternating operations: 1) local optimization and 2) global aggregation. In FeatureCloud, all running instances of a federated app have one of two roles (participant and coordinator) performing the respective federated operation. FeatureCloud expects precisely one coordinator and an arbitrary number of participants, leading to a star-based architecture.

After the local learning operation has been completed by a participant, it sends the local parameters to the coordinator. The coordinator collects these parameters and aggregates them into a collective (global) model, which is shared with the participants again. Depending on the type of ML algorithm, these two operations can alternate multiple times, e.g. until convergence or a pre-defined number of iterations has been reached (see Supplementary Figure S2). For some algorithms (e.g. random forest, linear regression), only one iteration is necessary. However, this strict separation between optimization and aggregation is not actively enforced by FeatureCloud. In many cases, aggregation can already start after the first parameters have been received, thereby increasing efficiency through parallelization of the computation. During implementation of a federated app, the distinction between coordinator and participant are of conceptual relevance. However, in practice, the coordinator can have local data as well that can be used for training. FeatureCloud therefore allows the coordinator to additionally assume the logical role of a participant.

Since FeatureCloud does not impose restrictions on the kinds of algorithms it supports, the running environment of the federated apps is kept very general. It allows for implementing any type of ML algorithm and an optional custom graphical user interface (GUI) for user interaction in form of a web-based frontend. This GUI can be used to receive input parameters, indicate the current progress or display the results. No direct Internet access is granted to the apps to avoid security risks.

## Federated algorithms

Since there are unique challenges for federating each algorithm, each machine learning model needs to be developed independently and therefore needs to be based on a different underlying federation mechanism. This means, each algorithm has its own challenges with regard to effectiveness, privacy, or scalability that need to be solved by the app developers.

### Linear and logistic regression

For the implementation of the linear and logistic regression apps, the methods introduced by Nasirigerdeh et al.[7] have been adapted from GWAS to a general machine learning use case. For linear regression, the local $X^TX$ and $X^TY$ matrices are computed by each participant individually, where X is the feature matrix and Y is the label vector. Then, they are sent to the coordinator, which aggregates the local matrices to global ones by adding them up. With these global matrices, the coordinator can calculate the beta vector identically to the non-federated method.

The logistic regression was implemented as an iterative approach. Based on the current beta vector, the local gradient and Hessian matrices of each participant are calculated and shared with the coordinator in each iteration. The coordinator aggregates the matrices again by adding them up, updates the beta vector, and broadcasts it back to the participants. This process is repeated until convergence or the maximum number of iterations (pre-specified for each execution) is reached.

Internally, the scikit-learn model API has been used to implement the apps[25,28]. In the performance evaluation, we used the default scikit-learn hyperparameters for the linear regression models. For logistic regression, penalty was set to none, the maximum number of iteration was set to 10.000 and the 'lbgs' solver was used to fit the models.

### Random forest

As an ensemble algorithm, random forest can be easily federated in a naive way. Our implementation trains multiple classification or regression decision trees on the local primary data of each participant. The fitted trees are then transmitted to the coordinator and merged into a global random forest. To account for different numbers of samples at each participant, each of them contributes a portion of the merged random forest proportional to their number of samples. To achieve a similar behavior as the centralized implementation, the size of the merged random forest is kept constant, meaning that an increasing number of participants decreases the number of required trees per participant. The federated computation happens in three steps, each involving data exchange: 1) participants indicate number of samples and receive total number of samples, 2) participants train required number of trees and the aggregator merges them into a global random forest, 3) participants receive the aggregated model to evaluate its performance on their data and share the results to obtain a global summary.

Since the aim is not achieving the highest possible accuracy, but comparing the federated to the non-federated version, the hyperparameters were set to the default values of sklearn, namely 100 decision trees, gini impurity minimization as splitting rule, and feature sampling equal to the square root of features. Pre-pruning parameters such as maximum depth, minimum samples per node and other constraints were not applied.

## Secure Multiparty Computation

One of the crucial aspects of FL is aggregating multiple local models from multiple participants. This leads to an imbalance of required trust. While every participant will be able to see the aggregated model after an aggregation step, only the coordinator knows all individual models. Interception of network traffic remains a threat for the same reason. To address this problem, the basic idea of SMPC is for each participant to split its local model into two pieces, a *masked* model (M - r), and the *mask* r, and send those pieces to two different entities. They can sum up all received pieces individually and then only exchange the sum. When adding the sum of masked models to the sum of all masks, the sum of local models can be obtained. In FeatureCloud, by design, there is only one aggregator. To still achieve the same behavior, the process is split into multiple stages.

Assuming n participants, each of them first creates a pair of public and private keys (e.g., using RSA), reveals its public key to all other participants through the coordinator. This allows each participant to prepare data destined for a particular participant by encrypting it with the respective public key, without the coordinator being able to read it. At the beginning of a training iteration, each participant first receives the global model. Each participant then creates an updated model using its local data as usual and masks the model with n different masks, one for each other participant, and encrypts each of those masks with the respective participant's public key. The masked model, together with the encrypted masks, is then sent back to the coordinator. The coordinator relays the encrypted masks to the participants who can decrypt their share of the masks and calculate the sum, which is then sent back to the coordinator. The coordinator finally sums up the masked models and the sums received from the participants to obtain the sum of local models. This achieves the same behavior of conventional SMPC, without the need of adding components or establishing additional communication channels. However, it leads to an increase in network traffic, growing quadratic with the number of participants. An illustration of this process can be found in Supplementary Figure S3.

## System architecture

FeatureCloud is a system consisting of several interacting parts, distributed between participants of a study and a central server. Figure 5 shows the system components and the communication channels between them. Further details about their implementation and employed technology can be found in Supplementary File A.

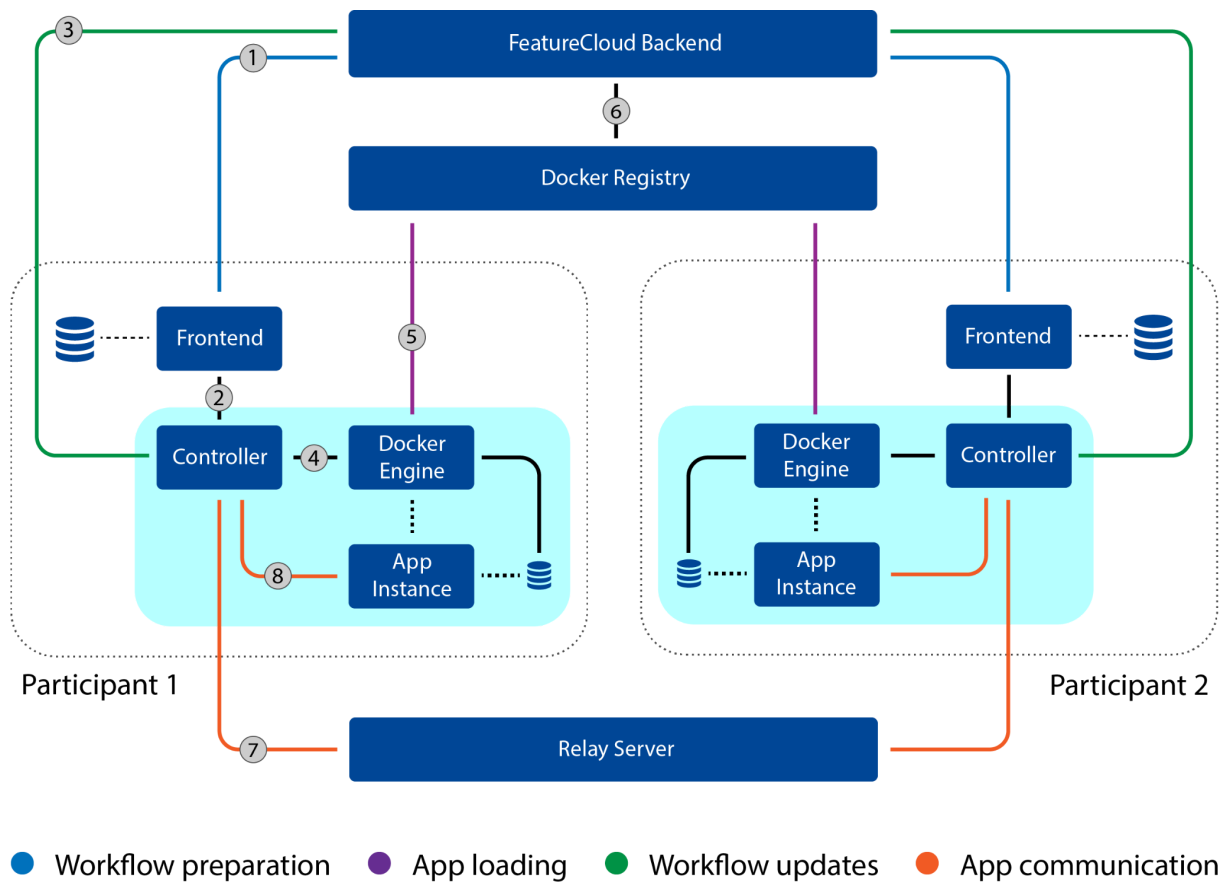

*Figure 5.* **System architecture of FeatureCloud with two participants.** *The Controller, Frontend, Docker Engine and App Instances are running locally at the participants. The FeatureCloud Backend and Docker Registry are running on FeatureCloud servers. The Relay Server can be run on a separate server or participants can use a provided instance from FeatureCloud. The components are connected via TCP/IP connections (straight lines). All links are HTTP-based, except for link 7, which uses a raw socket connection. Links 1 - 4 use JSON for serialization and links 4 - 6 use the Docker API.*

The *Frontend* is a web application running in a web browser. It uses the *FeatureCloud Backend* API (Figure 5, link 1) to offer all features of the AI Store and for collaborative project management. It is also connected to the *Controller* to allow for monitoring and handing over data for workflow runs (link 2).

The *Controller* is responsible for orchestrating the local part of the workflow execution. It receives information via the FeatureCloud Backend API (link 3), indicating which apps to execute next, and reports back about the progress. Contrary to the *Relay Server* traffic, this traffic only contains meta-information about the execution and no data used in the algorithms themselves. It uses the Docker API (link 4) to instruct the *Docker Engine* to manage containers which serve as isolated *App Instances* and pulls the images of the required apps for a workflow from the *Docker Registry* (link 5). When pushing new app versions, the Docker Registry ensures that the user is entitled to do so by verifying their credentials through the FeatureCloud Backend (link 6).

The *Relay Server* is responsible for transmitting all traffic of the federated algorithms via a secure socket connection (link 7). This central communication hub is aware of all participants and their roles in the federated execution and follows the required communication pattern, sending aggregated models to all participants and local model parameters to the coordinating party only. While FeatureCloud provides a relay server instance used by default, it is possible to use a private instance to completely shield the traffic from anyone outside the collaboration by adjusting the configuration file for the Controller.

Since apps are a dynamic system component, partly contributed by external developers, it is necessary to isolate their implementation. This is achieved by using Docker which ensures they cannot access system resources other than required, especially not the filesystem and network, and allows for putting limits on resource usage, such as CPU or memory. They receive their input data inside a Docker volume and communicate with the *Controller* through a defined API (link 8). This API is the main interface between externally developed apps and the FeatureCloud system. It is HTTP-based and requires the app to act as a web server, which means that it needs to wait for the controller to query for data and cannot actively send data by itself so that active network access can be forbidden.

# Acknowledgments

The FeatureCloud project has received funding from the European Union's Horizon 2020 research and innovation programme under grant agreement No 826078. This publication reflects only the authors' view and the European Commission is not responsible for any use that may be made of the information it contains. This work was further supported by the BMBF-funded de.NBI Cloud within the German Network for Bioinformatics Infrastructure (de.NBI) (031A532B, 031A533A, 031A533B, 031A534A, 031A535A, 031A537A, 031A537B, 031A537C, 031A537D, 031A538A).

# Supplementary Files

# A. Software Architecture and Implementation

This file contains information about technology, software architecture and implementation details for each of the integral FeatureCloud system components.

## Controller

The Controller is the local main part of the FeatureCloud system that is continuously run on a data holder's IT infrastructure. It needs to be able to handle large amounts of data and asynchronous tasks as well as keep up multiple socket connections and support HTTP-based and raw socket-based traffic. For this reason, this component has been implemented in Go (aka Golang), a native programming language developed for server applications. It allows for lightweight co-routines which are used to monitor app containers and regularly query for updates from the global FeatureCloud backend.

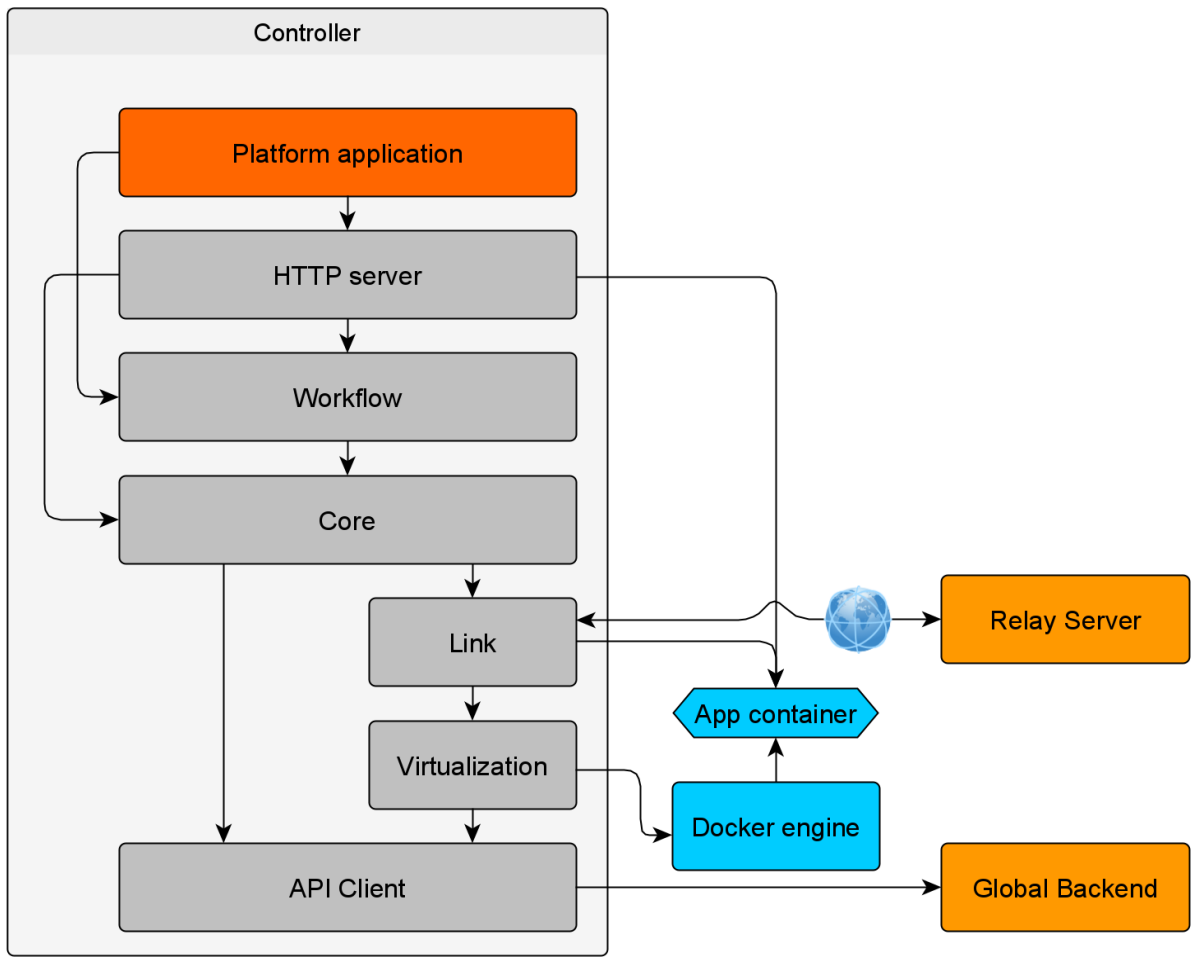

*Figure A1: Software architecture of the Controller. It uses a layered architecture preventing arbitrary access across layers by enforcing a partially ordered access hierarchy.*

The software architecture has a layered structure, with a decreasing level of abstraction from top to bottom (see Figure A1).

The platform application layer is the main entry point responsible for reading configuration values (e.g. local database credentials, address of the global backend) and starting an HTTP server and polling routines. The HTTP server provides endpoints for the frontend to control workflow-related tasks, such as loading data into the first input volume and showing container logs. It also relays traffic to the app-specific frontends. The workflow layer offers abstract functions for the HTTP server and takes care of workflow management, such as setting up and attaching volumes, starting containers, shutting them down, reacting to updates from the global backend (by using the data layer through the core layer). The core layer provides an abstraction of the core business logic, especially app container management and functions for testing apps during development. The link layer handles communication between app containers and the relay server, translating raw byte-traffic from the relay server to HTTP-based traffic for the containers and vice versa. The controller acts as an HTTP client in this case and the app containers as HTTP servers. This way, active access by the app containers to the Internet can be avoided. The virtualization layer is a direct abstraction of Docker, which allows for replacing the virtualization technique in the future if needed for security or compatibility reasons.

## Relay server

The relay server implements basic relay functionality for star-based federations of clients. It knows the role of each client (i.e. participant or coordinator) and treats their traffic accordingly. If data is received from a participant, it relays it to the coordinator. If it is received from the coordinator, it is broadcast to all clients. A relay server can handle multiple workflows at once. For that, it uses workflow-specific credentials chosen by the coordinator and automatically distributed to the participants by the global API. Like the controller, it is written in Go since it needs to efficiently handle large amounts of binary data, which Go is capable of.

## Global backend

The global backend mainly offers an HTTP API for controllers and the frontends. It is responsible for managing all necessary data related to projects, apps, users and data holders (sites). It is implemented in Django, a Python web framework that offers the functionality for this kind of task, particularly database abstraction, URL routing and web-related utilities (e.g. JSON serialization, HTTP abstraction).

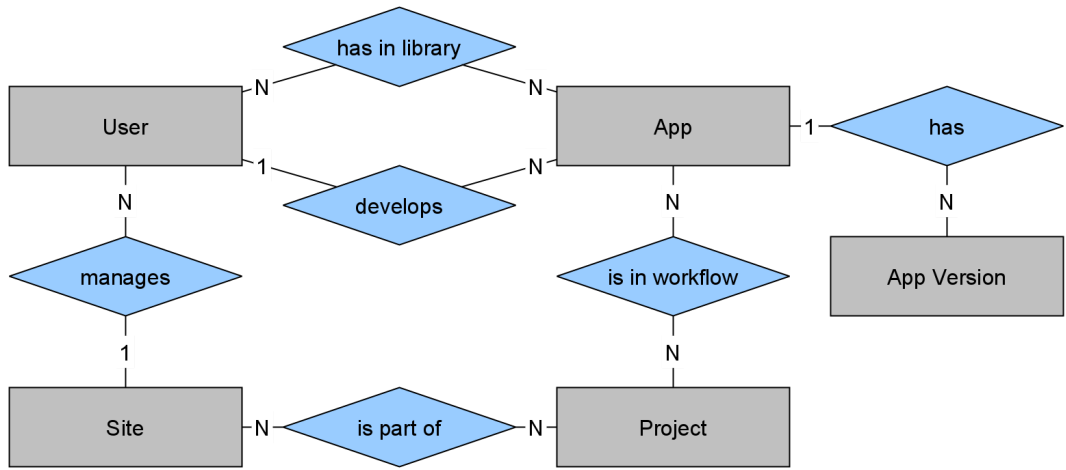

***Figure A2: E/R diagram of the data model in the backend.*** *Grey boxes represent entities, blue diamonds represent relationships.*

The E/R diagram of the data model is shown in Figure A2. The global backend allows controlled access to instances of these entities.

*User.* Each user has an email address and a hashed and salted password to log in to the FeatureCloud frontend, which then queries the global backend. In practice, a user is either a developer who has apps linked to them through the 'develops' relation or an end-user. Both, developers and end-users, can add apps to their library (relation 'has in library') and manage a site (relation 'manages').

*Site.* Sites have necessary contact information and represent data holder locations, e.g. hospitals or academic research institutions. Each site needs to run a controller instance to participate in projects (relation 'is part of'). When a site is part of a project, it can either assume the role of the coordinator or a participant.

*Project.* Projects encompass a workflow, descriptive information and a set of tokens allowing for joining a project. Tokens are not modeled explicitly. Instead, the 'is part of' table is used, which can have entries with a token string and where the related site is NULL. Once a site joins a project, this entry is linked accordingly and can no longer be used by anyone else.

*App.* Apps are AI applications that appear in the app store. They contain an image name, which needs to be used when pushing new versions of the app, an icon, a short and long description, tags, a category and a link to the source code. They are linked to a developer through the 'develops' relation and workflows they are part of through the 'is in workflow' relation.

*App Version.* New versions of apps are tracked automatically when pushing a new version via Docker by the developer and are linked to the respective app through the 'has' relation.

# Frontend

The frontend serves as a graphical user interface (GUI) for FeatureCloud users and developers. It is the only component FeatureCloud users directly interact with. It then calls the API of the controller or the global backend on behalf of the user, depending on the nature of the task. Since the frontend needs to be platform-independent, it has been implemented as a web application running inside a browser. This enforces a clear separation between GUI-related concerns and backend-related tasks by employing an HTTP-based API, as described earlier. Angular has been chosen as a web framework due to its high popularity, long-term support and extensive functionality.

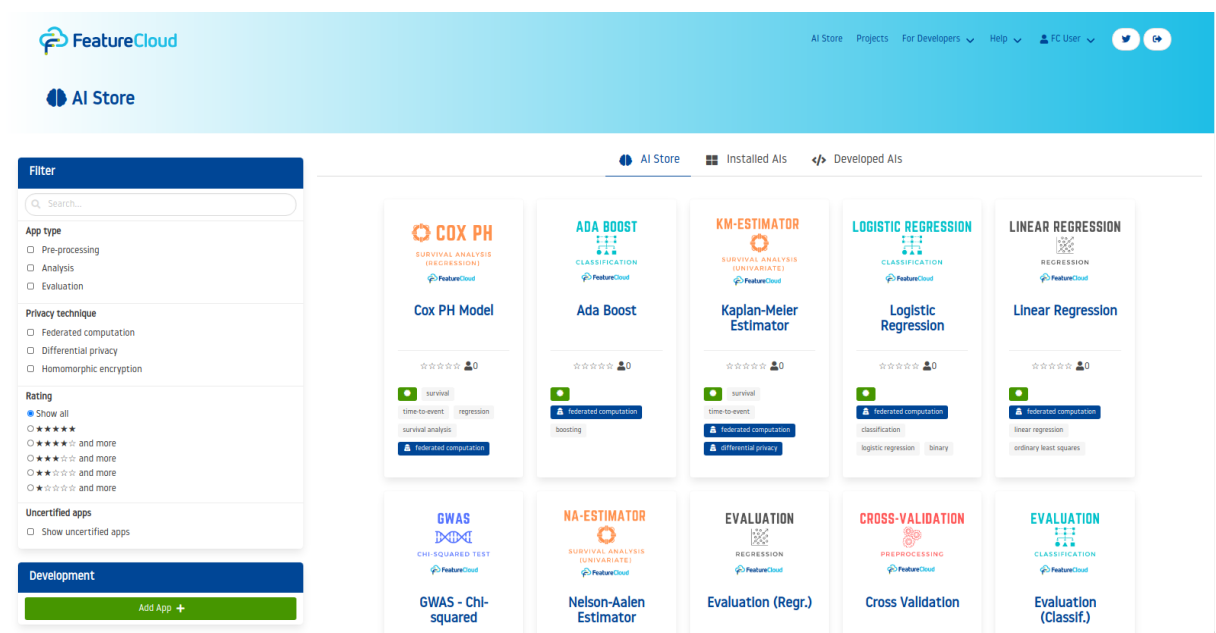

*Figure A3. The FeatureCloud AI Store.* *The FeatureCloud AI Store provides an overview about all the federated learning apps. While end-users can find the appropriate apps for their desired federated workflow here, app developers can contribute their own apps here. Once certified by the FeatureCloud team, they will be publicly available for all FeatureCloud users.*

The GUI is structured into the following sections accessible through the menu (see Figure A3): 1) Account management, 2) Site management, 3) App management, 4) Project management and 5) App testing, each divided into subsections again.

# Docker registry

The AI Store server is connected to the global backend that serves as an authentication server and a Docker registry. It performs two main tasks: relay queries from the local Docker engines using the Docker registry API[1] and protecting images from unpermitted access, in particular restricting pushing of images to the respective app developers. For that, the AI Store server provides endpoints to request a JWT token which is then attached automatically by the Docker CLI to authenticate consecutive actions. App developers need to be FeatureCloud users and use their FeatureCloud credentials to login. That way, the global backend acting as an auth server can check whether the user pushing an image is the corresponding app owner.

---
[1] https://docs.docker.com/registry/spec/api/

Like the controller, the relay server is written in Go for performance reasons. App images can be several GB large and pulling images is a task performed each time before a workflow step is executed, making performance a critical consideration.

# B. API Specification

A federated app acts as a web server polled by the FeatureCloud controller. Implementing the FeatureCloud API means implementing a web server that handles the following requests:

**POST /setup** When the participants are ready to start the federated execution (they are connected and prepared the input data) the platform will send the setup request. This is the starting point of the federated execution, the app can use it as a trigger to start the computation based on it's local data.

The request body contains the following information:

- `id` (string): the app instance identifier, determined by the platform
- `master` (boolean): this value specifies the role of the app instance: true for coordinator, false for participant
- `clients` (array of string): contains the identifiers of all participants

**GET /status** With the response to this request the federated app reports its current status. The app indicates if there is data to be transferred to the coordinator or if the execution of the app is finished.

The response should contain the following information:

- `available` (boolean): `true` if there is data to be transferred, otherwise `false`.
- `finished` (boolean): `true` if the app execution finished, otherwise false.
- `size` (int, optional): This value can be used to indicate the size of the data that will be transferred.

**GET /data** Using this API call apps can transfer the data to the platform.

The response body should contain the data to be transferred. If size was specified in the /status response, the platform will check if the content length matches the size value.
The platform reads the data and redirects in the following way, depending on the sender:

- If the data is coming from a participant, it will be redirected to the coordinator.
- If the data is coming from a coordinator, it will be broadcast to all other participants.

**POST /data** requests transfer of data from the app to the platform. The request body should contain the data to be transferred. The app should handle/consider the received data in the following way, depending on their role:
- If the receiver is a coordinator, the data is a packet from a participant (the ID of the sender is provided as a GET parameter 'client', e.g. /data?client=1)
- If the receiver is a participant, the data is a broadcast message from the coordinator

Besides implementing the above defined API, a federated app can have its own GUI, which is displayed by the FeatureCloud platform. The GUI is served by the app's web server, but its implementation is fully up to the app developer.

# C. Developer Tools

The development of algorithms involves intensive testing and debugging. For rapid development, it is crucial that these testing and debugging cycles are as quick as possible. Therefore, FeatureCloud comes with a local test framework, shown in Figure C1, that enables app developers to instantly run their application on their machine without deploying it first. When using this functionality, one has to specify the number of participants, i.e. app instances to simulate, and a data directory for each instance containing the respective input data. When started, the FeatureCloud controller creates one container for each instance and connects them logically identically on the developer's machine to a truly federated setup on different machines.

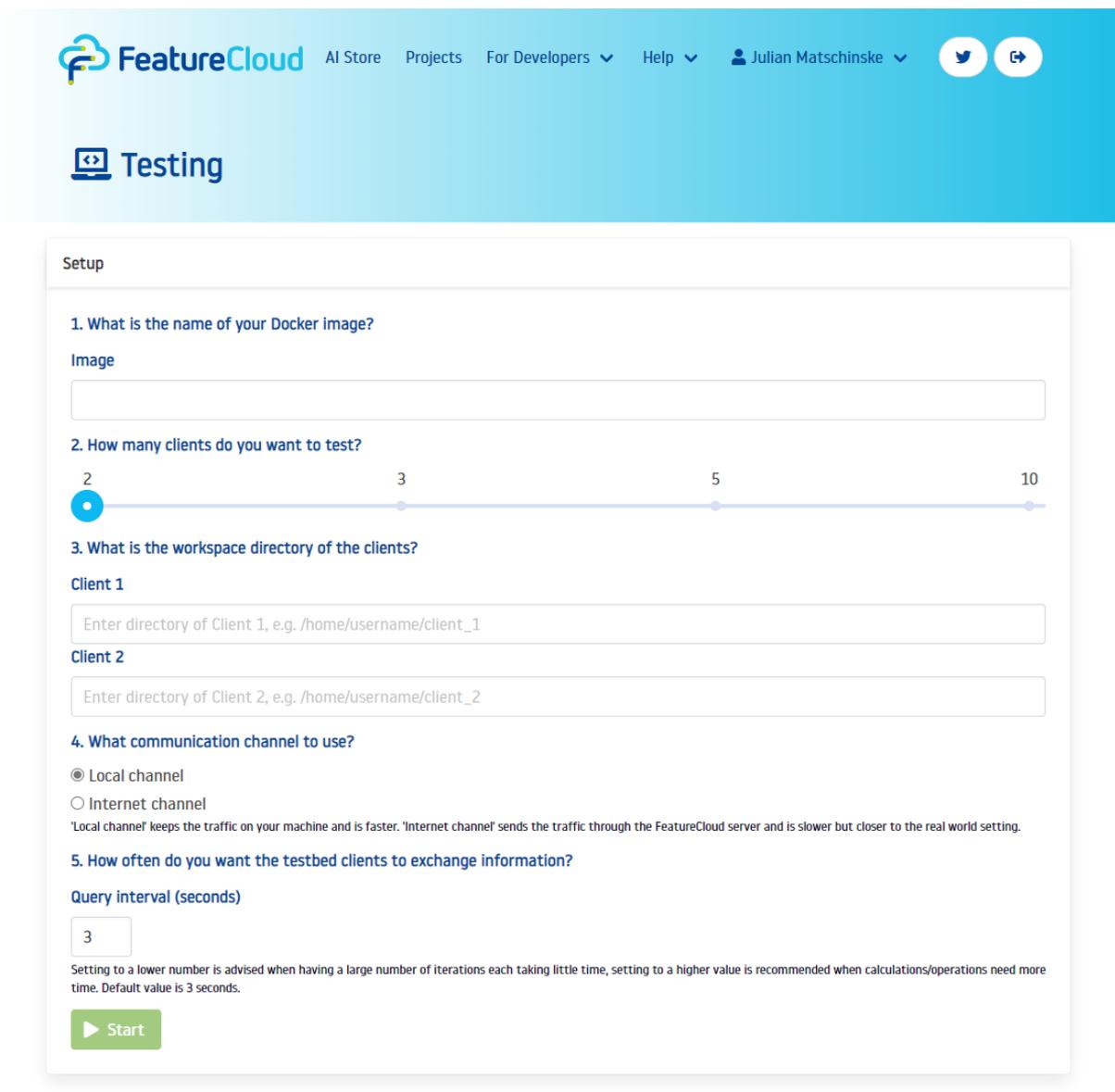

*Figure C1.* *Test test framework allows for specifying the Docker image, the number of participants to test, the directories containing test data, the type of relay channel (local or Internet) and the query interval used to fetch data from the running app containers.*

The API has deliberately been designed in an algorithm- and domain-agnostic way. This design leads to high flexibility but requires the app developer to implement all algorithm-specific functionality by themselves. To quickly introduce developers to the API and provide a convenient starting point for app development, FeatureCloud comes with a collection of easily extendable templates. This collection includes a minimal template with a demo Python implementation, stubs for all API calls and a blank demo frontend, and several federated implementations. They can be found on the FeatureCloud GitHub page[2].

---

[2] https://github.com/FeatureCloud

# D. Certification Process

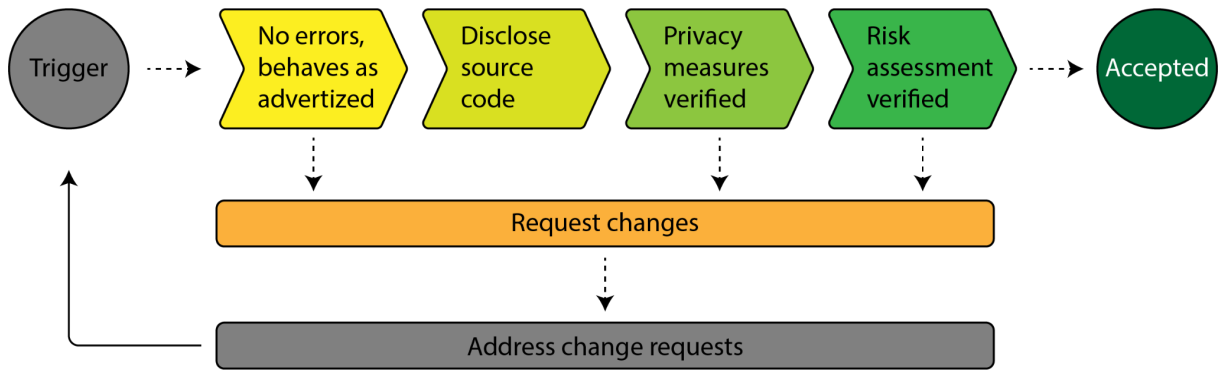

***Figure D1.*** *A new app and app updates need to undergo 4 stages during the certification process: Checking for errors, disclosing source code, verifying privacy measures, verifying risk assessment.*

All apps in the AI Store are isolated to the highest possible degree, i.e. they do not have access to the filesystem or the Internet. On top of that, the apps need to abide by our privacy standards. Herefore, they must undergo the following steps before being certified by a member of the FeatureCloud consortium and thus showing up in the public AI Store (see Figure D1).

1. The app needs to run without errors and provide the features advertised in the app description.
2. The complete source code must be disclosed so that the app image can be reproduced.
3. Privacy mechanisms that have been declared (e.g. DP, HE) need to be implemented correctly and as stated in the app description.
4. The app developer needs to elaborate on the data the app sends to the coordinator and assess the level of potential privacy leakage risks. In particular, no raw data must be handed over to the coordinator under any circumstances.

If these criteria are fulfilled, the app is built by the certifying party and pushed to the FeatureCloud AI Store. This way, it is ensured that no other image than the certified one will be executed on the systems of the end-users. The process will be repeated for each update of the app.

# S. Supplementary Figures

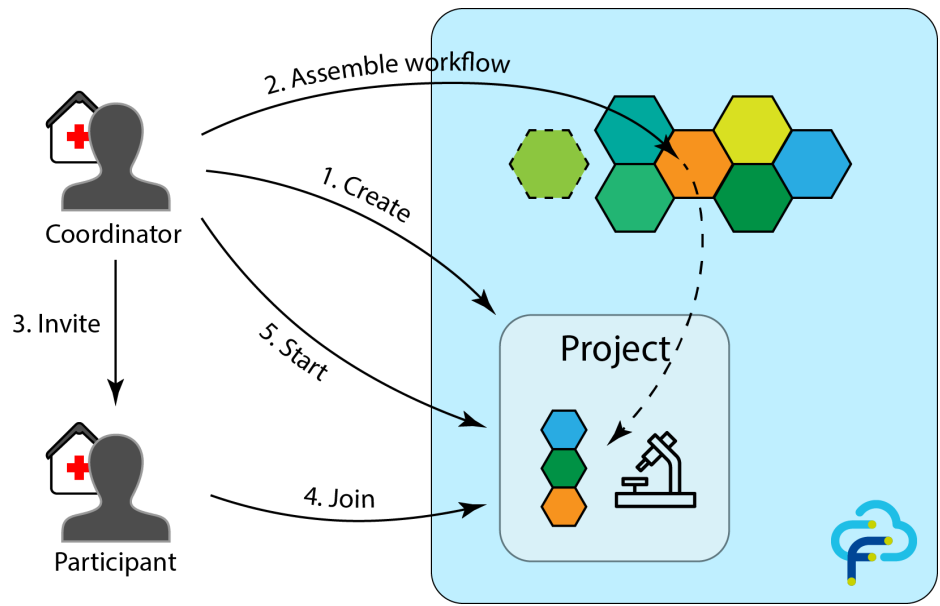

*Figure S1. Creating and setting up a federated project.* One of the partners takes the role of the coordinator and creates a new project on the website (1). After that, the coordinator defines the workflow by adding the apps to the project's workflow they want to run collectively (2). Then, the other partners are invited by sending a randomly generated token to each of them (3), which is unique and allows for joining the project (4). When all partners have joined, the coordinator triggers the execution on the FeatureCloud website and the workflow begins to run (5). During workflow execution, active interaction with the end-user can be required, depending on the apps.

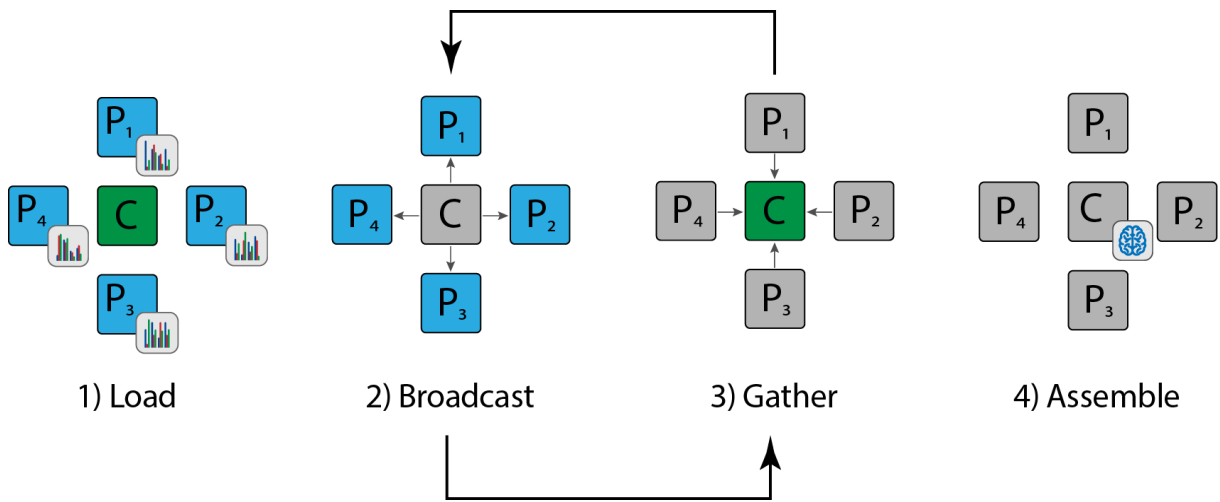

*Figure S2. Four stages of federated execution in FeatureCloud.* The four main stages are 1) local data loading, 2) broadcasting a global model, 3) gathering local models, 4) assembling results. Stage 2 and 3 can be repeated depending on the executed algorithm. 'C' and 'P' stand for coordinator and participant, respectively.

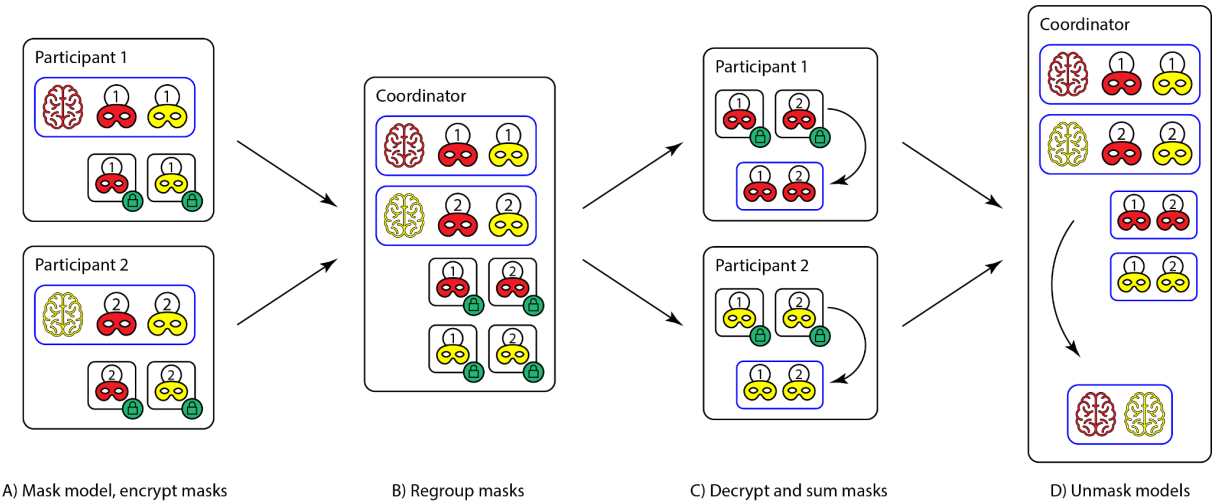

*Figure S3. Two participants and the coordinator create a global model without either seeing local models of another party.* Blue rectangles represent the sum of their content, blue locks represent encrypted data. The color of the models show their origin (red for participant 1, yellow for participant 2), the color of the masks indicate their destination and with whose public key they were encrypted.